\documentclass[11pt]{article}
\usepackage[final]{acl}
\usepackage{times}
\usepackage{latexsym}
\usepackage[T1]{fontenc}
\usepackage[utf8]{inputenc}
\usepackage{microtype}
\usepackage{inconsolata}
\usepackage{graphicx}
\usepackage{booktabs}
\usepackage{array}
\usepackage{amsmath}
\usepackage{amssymb}
\usepackage{algorithm}
\usepackage{algpseudocode}
\usepackage{url}
\usepackage{float}

\title{Knowing Which, Not How Much: \\ Language Models Order Party Positions They Cannot Place}

\author{Yousef Attia \and Jaros{\l}aw A. Chudziak \\
  Institute of Computer Science \\
  Faculty of Electronics and Information Technology \\
  Warsaw University of Technology \\
  Warsaw, Poland \\
  \texttt{\{yousef.attia.stud, jaroslaw.chudziak\}@pw.edu.pl}}

\begin{document}
\maketitle

\begin{abstract}
Language models are increasingly asked where real actors stand: what a party's platform prioritises, what an institution has committed to. Such judgments are gradable, carrying both an order and a magnitude. Nearly every benchmark that tests them asks about one entity at a time, then reads that score as the model's knowledge of the domain.
Order and magnitude, however, need not be acquired together. Text about social entities carries far more comparison than measurement, so a model may hold a dependable ordering with no anchored scale beneath it.
We ask whether a model can \emph{place} a documented social position on a scale or only \emph{order} the actors it compares, and what that means for how such knowledge is measured.
We test this on party positions in Japan and South Korea, where the Comparative Manifesto Project's coded data supplies a rare external answer key.
Asked to order four parties, models beat the strongest fixed-answer baseline by more than a factor of two; asked to place one party on the same scale, they fall below it. The ordering survives inside those failed placements, and supplying the coded values lifts placement to near ceiling.
Comparative and absolute questions about the same fact therefore measure different things here, and for party issue salience a benchmark built from one cannot speak for the other.
\end{abstract}

\section{Introduction}

As elections crowd the global calendar, voters, journalists and civic-education tools reach for a conversational system to ask where a party stands, and such systems are already a default first stop \citep{luettgau2025conversational, kuznetsova2024generative}. A question like \emph{how far does this party prioritise welfare} is \emph{gradable}, and its two halves come apart: which parties rank above which, and how high up the scale any one of them sits. Benchmarks for questions of this kind ask about one actor per prompt, and the accuracy they report is then treated as a measure of subject-matter competence. That reading also decides where evaluation is possible at all, because building the answer key is the expensive part, so coverage follows the funding of the key more than the reach of the model.

Order and magnitude need not be acquired together. Human judgment has long been known to be steadier under comparison than under absolute rating \citep{thurstone1927law, verhavert2019meta}, an asymmetry so well established that preference learning collects pairwise choices over scores, and text about social entities inherits the same shape. A party is described as more welfare-oriented than its rivals far more often than its manifesto is quantified. A model trained on that text can therefore learn the ranking without ever learning the quantities under it, and when an absolute judgment is demanded it cannot report that the level is missing, only invent one. What a model appears to know would then depend on which quantity a question asks for, an order or a level, more than on how it is worded. If so, the prevailing single-item format measures the weaker of the two capacities and reports it as subject-matter competence.

\textit{This leads to our key question: can a model \emph{place} a documented social position on a scale, or only \emph{order} the actors it is given to compare, and what follows for how such knowledge is measured?} Party positions are the test case for this question; the question itself is about measurement.

We make this testable through three claims. All three are tested on the same benchmark: eight parties and ten policy issues across Japan and South Korea, answered by seven production models in two framings.

\emph{First, absolute placement: a model can place a named party on the four-level emphasis scale.} We would call this reliable only if exact-match accuracy cleared the majority-class rate a constant answer already achieves.

\emph{Second, relative ordering: asked to compare the same four parties on the same issue, a model names the highest- and lowest-emphasis party.} We count this reliable only above the rate the best single fixed guess achieves.

\emph{Third, any gap between the two comes from missing knowledge, with the scale itself usable.} We would accept this only if supplying the coded values both lifts placement sharply on the same scale and repairs the model's disagreement with the ordering implied by its own placements.

We report each claim against its criterion. The paper makes two contributions. The first is evidence that comparative and absolute questions about the same documented fact measure different capacities, so that, for party issue salience at least, a benchmark built from one framing says little about the other. The second is a benchmark for cross-party attribution on professionally coded manifesto data from two Asia-Pacific democracies, released with notes on two pitfalls for reuse: one CMP category behaves opposite to its name, and the reference-free consistency metric is sensitive to how verdicts are read out. The agreement check inside the third claim consults no answer key, which is what lets it travel to the many settings that have none.\footnote{Code and data: \url{https://github.com/Deeperr0/cross-party-attribution-benchmark}.}

\section{Background and Related Work}

\textbf{Comparison versus absolute rating.} That comparative judgment is steadier than absolute rating is old and well replicated in humans, from the law of comparative judgment \citep{thurstone1927law} through meta-analyses of its reliability \citep{verhavert2019meta} and domain studies where pairwise protocols lift interobserver agreement over Likert scales \citep{hoeijmakers2024pairwise}. The finding carries over to language models used as evaluators, which are more reliable comparing two candidates than scoring each alone \citep{liusie2024comparative, qin2023pairwise}, and the same holds when the object is a scalar social-science construct \citep{licht2025scalar}. That literature's object is almost always the quality of a supplied text, judged against human preference with no external key, so ``comparison is easier'' cannot be separated from ``comparison aligns better with the reference''; and its remedy is a readout fix, since pointwise scoring recovers much of the gap given a sufficiently fine ordinal scale \mbox{\citep{godfrey2025likert}} or a probability-weighted decode \citep{licht2025scalar}. Our setting differs in both respects: nothing is supplied for the model to judge, the key is external and pre-existing, and the deficit we find is repaired by supplying the fact, where a finer readout leaves it in place.

\textbf{What parametric knowledge is reachable.} A model can encode a fact and still fail to retrieve it. \citet{calderon2026recall} report that frontier models encode 95--98\% of the facts they test yet fail to recall 25--33\% of them without inference-time computation. The sharpest case is directional: asked a learned relation backwards, a model that cannot \emph{produce} the answer will still \emph{pick} it out of a supplied list, an asymmetry \citet{berglund2023reversal} report independently. The fact is held, and only the route to stating it is blocked. That is the main alternative explanation for any comparative-versus-absolute gap. Our design controls for it: both framings supply the alternatives in the prompt (four emphasis levels, or four named parties), so recognition is available in both conditions, yet placement still fails. The two framings differ in what is being identified: a level in one, a party in the other.

Where a magnitude exists in text, a model can hold it and route around it: \citet{lehmann2026shortcut} compare a model's pairwise entity judgments against the ordering implied by its own predicted attribute values and find the implied ordering substantially \emph{more} accurate, with popularity and mention-order heuristics overriding numbers the model demonstrably has. Their attributes are quantities written down as numbers, such as river lengths and city populations drawn from Wikidata, so a cardinal value exists to be bypassed; ours is a coded salience no one records. Their intervention is chain-of-thought, a reasoning fix, where ours is fact injection, a knowledge fix.

\textbf{Political attribution as the testbed.} Language models are now routinely used to estimate party positions, and the picture is consistent. Placements from party names alone track expert surveys closely but agree least with manifesto coding \citep{dileo2025mapping}, estimated positions compress the ideological extremes that human coders record \citep{caliskan2025generative}, and issue \emph{salience} is recovered less well than issue position \citep{benoit2026salience}. Related work also measures the model's own leanings \citep{rozado2024political, motoki2024more, hartmann2023political, ceron2024beyond, buyl2024large} or conditions models on personas to simulate electorates \citep{kang2026dynamok}. \citet{batzner2024germanpartiesqa} benchmark commercial models against German parties' positions, and \citet{gudino2024augmented} find that models' political choices shift with the order in which options are presented. A neighbouring strand studies political \emph{language} itself, using language agents to study affective polarisation \citep{malvicini2026affective} and reader interpretation of rhetorical appeal \citep{gajewska2026ethos}. That work shares our concern with what a model reports about political content, but it measures reception and stance where we measure fidelity to a documented record.

Our object is narrower than any of these. We ask whether a model can report what a named party documented, and whether that ability depends on being asked comparatively. Where the studies above supply the text to be scored or target the model's own stance, ours withholds the source entirely.

\textbf{Ground truth source.} The CMP/MARPOR corpus is established in political science for scaling positions and auditing coding methods against human coders. We repurpose that coded data as an answer key for scoring model-generated position statements, which is what lets the key-free agreement measure be validated against a standard before it is used where none exists. Appendix~\ref{app:related} tabulates how the closest prior work differs in task, ground truth, and region. The design below is built to give a case where the model must answer from what it already holds and the answer can still be checked.

\section{Data and Benchmark Construction}

Two things are required of the benchmark. It needs real, checkable ground truth for each party's position, and it needs issues chosen so that some separate the parties and some do not. This section describes both, along with the labelling choices that turn coded manifesto percentages into ordinal targets.

\subsection{Ground Truth and Party Selection}

We use the Manifesto Project (CMP/MARPOR) main dataset \citep{manifesto2025}, in which trained coders segment each party's election manifesto into quasi-sentences, assign every segment one of a fixed set of policy categories (\texttt{perXXX}), and report the percentage of the manifesto falling in each. The measure is therefore \emph{salience}, how much a party wrote about an issue, not its stance on it. The dataset covers sixty-seven countries; Japan and South Korea were chosen for that coverage alone, as the only non-Western Asian democracies coded across enough elections to support this comparison. A corpus with different coverage would have pointed the same method at different countries, which is the coverage problem in concrete form.

We select the four largest, most internationally recognisable parties from each country's most recent election in the dataset (Japan 2021: Liberal Democratic Party, Japanese Communist Party, Constitutional Democratic Party of Japan, Komeito; South Korea 2020: Democratic Party of Korea, United Future Party, Justice Party, People Party). This follows from a feasibility pilot. Models have thin, unreliable knowledge of minor parties regardless of question format, which would conflate a knowledge gap with the effect under study.

\subsection{Issue Selection}

We select ten policy categories spanning a range of real cross-party variance, checked in the data and not assumed from the category's English name: six (\texttt{per403} Market Regulation, \texttt{per411} Technology/Infrastructure, \texttt{per402} Incentives, \texttt{per501} Environmental Protection, \texttt{per104} Military, \texttt{per303} Governmental/Administrative Efficiency) show genuine cross-party spread, while four show none. Three of these are low-emphasis anchors where every party clusters near zero (\texttt{per408} Economic Goals, \texttt{per704} Middle Class/Professional Groups, \texttt{per406} Protectionism); the fourth, \texttt{per504} Welfare State Expansion, is the mirror case where all cluster high.

\textbf{A category-definition hazard.} One selected category, \texttt{per408} (``Economic Goals''), is defined in the CMP codebook as a residual class, meaning general economic statements not assigned to any more specific category, with specific positions explicitly overriding it. A party with a detailed economic platform therefore scores \emph{low} precisely because its specifics are coded elsewhere, the reverse of what the English name suggests, and models consistently and confidently rate it a ``core priority'' for such parties. We exclude \texttt{per408} from all headline results (roughly 10\% of the dataset) and report it separately. A category-definition mismatch between a benchmark's labels and its underlying coding scheme is a live hazard for evaluation built on secondary-coded social-science data.

\subsection{Gold Labels and Ties}
\label{sec:gold}

Raw CMP percentages are converted into four ordinal buckets (not a stated priority, $<$1\%; minor emphasis, 1--5\%; moderate emphasis, 5--10\%; major emphasis/core priority, $>$10\%). Formally, a raw manifesto percentage $c$ maps to
\begin{equation}
b(c) = \begin{cases} 0 & c < 1\% \\ 1 & 1\% \le c < 5\% \\ 2 & 5\% \le c < 10\% \\ 3 & c \ge 10\% \end{cases}
\label{eq:bucket}
\end{equation}
Bucketing is deliberately lossy. It discards distance within a band, which makes exact-match scoring possible at the price of scoring a near miss and a wild miss alike. Section~\ref{sec:results} therefore reports placement under both exact and adjacent-credit scoring, each against its own baseline.

For a country's party set $P$ on issue $j$ with true percentages $c_{p,j}$, the gold top set is the tie-aware arg-max $T_j = \{p \in P : c_{p,j} = \max_{p'} c_{p',j}\}$ and the gold bottom set $L_j$ the analogous arg-min. We record the full tied set, so a model is charged only with errors the data itself settles.

\begin{figure*}[t]
\centering
\includegraphics[width=\textwidth]{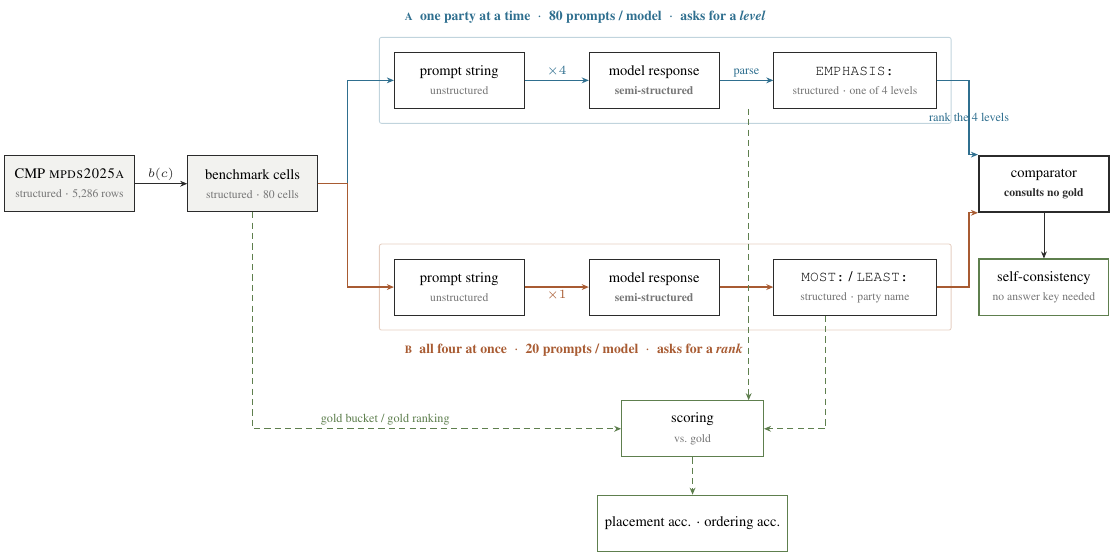}
\caption{The life cycle of the data. Coded CMP percentages are quantised into
ordinal buckets, giving each benchmark cell a gold level and a tie-aware gold
ranking; the same cell is then put to a model two ways. The two lanes are separate
API calls with no shared history, and both ask a four-alternative question: four
emphasis levels in \textsc{a}, four named parties in \textsc{b}. Each lane starts
from a structured prompt; the model answers in free prose that ends with one
machine-readable tagged line (Section~\ref{sec:readout}), and both lanes are parsed
identically. Dashed paths are the only ones that read the gold column.}
\label{fig:pipeline}
\end{figure*}

\section{Experimental Setup}

With the benchmark fixed, the experiment varies only how each party-issue pair is asked. This section states which models we query, how the two framings are posed, how answers are read out, and the grounding intervention that separates missing knowledge from an unusable scale.

\subsection{Models and Prompts}

We evaluate seven language models spanning open, regional, and frontier providers: Cohere Command-A, two Groq-hosted \texttt{gpt-oss} models (20B and 120B parameters, allowing a small-versus-large comparison within one family), Mistral Large, Anthropic Claude Sonnet, Google Gemini 2.5 Flash, and DeepSeek-V3. These are current, general-purpose production assistants, the same class of system a voter reaches for, queried through their APIs and used plainly, with a single neutral system prompt (``You are a knowledgeable political analyst. Answer concisely and directly. If you are not confident you know a party's real platform, say so explicitly rather than guessing.'') followed by the question, at temperature~0 where the provider supports it (the Claude Sonnet generation deprecates the parameter and is queried at its default). We assign no persona, attach no retrieval, and do no fine-tuning, so the model answers from its own parametric knowledge; the grounding intervention of Section~\ref{sec:grounding} is the single, diagnostic exception. Each of the 100 prompts is a separate API call with no conversation history, so each framing is answered without sight of the other, and any agreement between them comes from the model itself. Figure~\ref{fig:pipeline} traces the whole chain, from the coded percentages through both framings to the four reported quantities.

Each party-issue pair is queried in two conditions. In the \textbf{absolute} condition the model is asked about one named party's stated position and its relative emphasis (a core priority, a moderate focus, a minor mention, or not really addressed). In the \textbf{comparative} condition it is asked to compare all four of a country's parties on the same issue in one prompt and name which places the most and least emphasis on it. Party names are country-qualified, and we substitute each party's commonly recognised English name for the CMP dataset's literal translation while retaining the CMP name for citation. The input is a prompt string and nothing more: no persona, no retrieval, no metadata. Each model answers the same 100 prompts, 80 absolute (eight parties $\times$ ten issues) and 20 comparative (two countries $\times$ ten issues).

\subsection{Reading Out the Verdict}
\label{sec:readout}

Models answer in free prose, and the verdict must be recovered from it. We ask each model to answer as it otherwise would and then append a final tagged line: \texttt{EMPHASIS:} one of the four offered labels or \texttt{UNKNOWN} in the absolute condition, \texttt{MOST:}/\texttt{LEAST:} a party name in the comparative one. The prose task is unchanged, so the model answers as it would in normal use; \texttt{UNKNOWN} lets a model decline to guess; and the labels are the words already in the question stem, introducing no new vocabulary. Tag compliance was 96.7\% overall (77\% for one model, 100\% for the rest); the few untagged responses fall back to a keyword classifier, counted and reported.

Scoring then compares the recovered bucket to the CMP gold value. A response is scored correct, one bucket off (partial), two or more buckets off and equal to a rival party's true bucket (rival match), two or more buckets off and equal to no party's bucket (no match), or an explicit statement of uncertainty (\texttt{no\_knowledge}). No automated judge model is used anywhere in the pipeline.

Formally, placement is correct when $\hat{b}_{m,p,j} = b(c_{p,j})$, where $b(c_{p,j})$ is the coded bucket for party $p$ on issue $j$ (Equation~\ref{eq:bucket}) and $\hat{b}_{m,p,j}$ is the bucket model $m$ states for it, restricted to the absolute framing as $\hat{b}^{\mathrm{abs}}$. The agreement half of the third claim is measured as self-consistency on the highest-emphasis party,
\begin{equation}
\mathrm{SC}^{\mathrm{hi}} = \frac{1}{|\mathcal{C}|} \sum_{(m,j) \in \mathcal{C}} \mathbb{1}\!\left[\, s_{m,j} \in \tau\bigl(r^{\mathrm{hi}}_{m,j}\bigr) \,\right],
\label{eq:sc}
\end{equation}
taken over the cases $\mathcal{C}$ in which the absolute answers actually distinguish parties. There $s_{m,j} = \arg\max_{p} \hat{b}^{\mathrm{abs}}_{m,p,j}$ is the party those answers rank highest, $r^{\mathrm{hi}}_{m,j}$ is the party the comparative answer names highest, and $\tau(\cdot)$ is the model's own tie group. Equation~\ref{eq:sc} compares two of the model's own outputs and never consults $b(c)$. Cases where fewer than two parties are scoreable, or where all scoreable parties land in one bucket, are dropped, since there is no ordering to be consistent about.

\subsection{Grounding Intervention}
\label{sec:grounding}

To separate missing knowledge from an unusable scale, we rerun every prompt with one change: a short, neutrally-worded prefix stating all four parties' real coded bucket for that issue, party order shuffled per issue to avoid positional priming. The question that follows is identical, and responses are read out and scored identically. This grounding is strong by design: the prompt states the answer outright, with no retrieval step. Because grounding changes only what the prompt contains and nothing about how the answer is read, whatever it moves is attributable to the facts being present.

\section{Results}
\label{sec:results}

\begin{table}[t]
\centering
\footnotesize
\setlength{\tabcolsep}{3pt}
\begin{tabular}{lccccc}
\toprule
\textbf{Model} & \textbf{Corr.} & \textbf{Part.} & \textbf{Rival} & \textbf{None} & \textbf{NoKn.} \\
\midrule
DeepSeek-V3 & 26.4 & 44.4 & 4.2 & 20.8 & 4.2 \\
Gemini-Flash & 25.5 & 30.9 & 7.3 & 36.4 & 0.0 \\
gpt-oss-120b & 25.0 & 38.9 & 2.8 & 20.8 & 12.5 \\
Cohere-A & 22.2 & 31.9 & 5.6 & 25.0 & 15.3 \\
gpt-oss-20b & 21.1 & 35.2 & 5.6 & 32.4 & 5.6 \\
Claude-Sonnet & 20.8 & 22.2 & 0.0 & 2.8 & 54.2 \\
Mistral-L & 16.7 & 50.0 & 5.6 & 26.4 & 1.4 \\
\midrule
Overall (n=486) & 22.4 & 36.4 & 4.3 & 23.0 & 13.8 \\
\bottomrule
\end{tabular}
\caption{Absolute-placement verdicts by model (\%), sorted by accuracy. Corr.=Correct, Part.=one bucket off, Rival/None=two or more buckets off and matching a rival's bucket / no party's bucket, NoKn.=No knowledge. Models differ far more in calibration than in accuracy: Claude Sonnet declines on over half of items while others rarely do ($\chi^2=126.7$, $p<10^{-24}$, Cram\'er's $V=0.51$).}
\label{tab:singleparty}
\end{table}

\begin{table}[t]
\centering
\small
\setlength{\tabcolsep}{4pt}
\begin{tabular}{lccc}
\toprule
\textbf{Task and scoring} & \textbf{Model} & \textbf{Base} & \textbf{Ratio} \\
\midrule
Order four parties & 58.1 & 22.2 & \textbf{2.62}$\times$ \\
Place one, exact & 22.4 & 37.9 & \textbf{0.59}$\times$ \\
Place one, $\pm$1 bucket & 58.8 & 80.7 & \textbf{0.73}$\times$ \\
\bottomrule
\end{tabular}
\caption{The dissociation. ``Base'' is the best single constant answer under that task's own scoring rule: always naming one fixed party (ordering) or always giving one fixed bucket (placement). The two 58\% figures sit on opposite sides of their baselines, which is why raw accuracies must not be compared directly.}
\label{tab:dissociation}
\end{table}

\textbf{Models order parties well above baseline and place them below it.} Asked to name the highest-emphasis party among four, models are correct 58.1\% of the time (n=117) against the 22.2\% a fixed guess of the single most frequent gold winner would achieve, and 48.3\% on the lowest (n=116) against 27.8\%. Both beat the rate a random pick achieves (25\% for the highest party, which is never tied; 35.3\% for the lowest, where 24 cells hold a three-way tie; binomial test, Appendix~\ref{app:holm}). Asked instead to place one named party on the same four-level scale, they are exactly right 22.4\% of the time (n=486, Table~\ref{tab:singleparty}), \emph{below} the 37.9\% that answering ``not a stated priority'' to every question would score.

Crediting adjacent buckets raises placement to 58.8\%, but the best constant answer under that rule scores 80.7\%, so the model still loses to a fixed reply. Both scoring rules put ordering at more than twice its baseline and placement below its own (Table~\ref{tab:dissociation}). The spread across models is narrow and includes the frontier systems. The gap is widest on the six variant issues, where ordering the highest party reaches 69.6\% against 25\% chance while placement falls to 21.9\% against a 41.9\% constant; on the three anchor issues, where all parties sit at one level, ordering the highest party is not significant (34.2\%; Appendix~\ref{app:split}).

\textbf{The ordering survives inside the answers whose levels are wrong.} Placement and ordering were asked in separate sessions, so we also derive an ordering from the model's own placements. Ranking the four parties by the buckets the model itself assigned yields an ordering that matches the gold extreme 52.8\% of the time (n=36, counting only cases where those buckets name a single party), against 60.0\% for the directly-elicited comparison on the same cells, a difference well within chance (Fisher's exact test, Appendix~\ref{app:holm}). The placements those orderings are built from are correct 24.3\% of the time on those same cells. Both routes to an ordering therefore work about equally well, while the levels they are built from fail. The levels are displaced, and displaced consistently enough in one direction that the ranking they induce survives.

\textbf{The failure has a direction.} The levels models state run high. Across the same 486 items they call an issue a moderate or core priority 72.2\% of the time when the coded truth is 33.7\%, and answer ``not a stated priority'' on a small fraction of the items where it is the single most common correct answer. Answers two or more buckets off match a rival party's true bucket 15.8\% of the time (n=133), below the 19.2\% expected if those answers fell at random, so we read rival matches as coincidence and draw no conclusion about cross-party contamination. The error is therefore a systematic upward shift, more than noise around a correct level.

\textbf{Grounding shows the scale is answerable.} Handed the four coded buckets in the prompt, placement accuracy rises from 22.4\% to 95.6\% on the same scale, the same prompt, and the same read-out procedure. The four-level scheme is therefore learnable, the question answerable, and the scorer capable of recognising a correct answer. This is consistent with the ungrounded failure being missing knowledge, though it does not exclude a failure to map a held belief onto the scale (see Limitations).

\begin{table}[t]
\centering
\small
\setlength{\tabcolsep}{4pt}
\begin{tabular}{lccc}
\toprule
\textbf{Measure} & \textbf{Ungr.} & \textbf{Gr.} & \textbf{Sig.} \\
\midrule
Absolute placement & 22.4 & 95.6 & --- \\
Self-consistency (highest) & 86.2 & 96.8 & $p=0.039$ \\
Self-consistency (lowest) & 77.5 & 96.9 & $p=0.0006$ \\
\bottomrule
\end{tabular}
\caption{Effect of grounding. Supplying the coded values lifts placement to near ceiling and significantly raises agreement between a model's comparative answer and the ordering implied by its own absolute answers (Fisher's exact). Self-consistency $n$: 80/80 ungrounded, 63/64 grounded.}
\label{tab:grounding}
\end{table}

\textbf{Self-contradiction is modest, and grounding reduces it.} Where a model's own absolute answers assign different parties to different levels, its separately-elicited comparative answer names the same party as highest 86.2\% of the time (95\% Wilson CI [77.0, 92.1], n=80) and the same lowest 77.5\% ([67.2, 85.3], n=80). Models do not differ significantly in this agreement (Appendix~\ref{app:holm}). Agreement on the highest party is high enough that we cannot call the models self-contradictory on it; agreement on the lowest party is clearly weaker. Supplying the coded values raises agreement to 96.8\% and 96.9\% (Table~\ref{tab:grounding}). After Holm correction across the nine tests we report, only the lowest-party improvement clears its threshold (Appendix~\ref{app:holm}). \textbf{The third claim therefore holds for the lowest party and is directionally consistent for the highest}: if the contradiction were an independent defect of the comparison mechanism, handing over the facts would not repair it. On the lowest party it does, which is consistent with that instability following from the missing level.

\textbf{How the verdict is read out changes the measured contradiction by twenty-eight points.} The comparative verdict can also be recovered from prose instead of from the tagged line, by locating a superlative phrase and taking the nearest party name in a surrounding character window, which is how a metric of this kind is computed when no tagged line is available. On the same prompts, run without the tagged line, that procedure reports self-consistency of 58.0\% against the 86.2\% recovered from the tagged line (64.0\% versus 77.5\% on the lowest party; tests in Appendix~\ref{app:holm}), and it resolved most and least to the same party, a logically impossible answer, on 24.6\% of responses. The grounded condition bounds the error directly, because there the correct answer is in the prompt and we know what the model should say. Prose extraction recovers it 72.3\% of the time, the tagged line 95.6\%. Extraction error of that size lands on both sides of a self-comparison and inflates apparent disagreement. Absolute-placement accuracy is essentially unchanged between the two read-outs (20.7\% versus 22.4\%), so the models answered comparably and the difference lies in the measurement. We report this because reference-free consistency metrics are widely used and, on this evidence, their headline value can be dominated by the extraction step that precedes them.

\section{Discussion}

The two framings ask for the same fact and get different competence. One reading of this pattern is that what a model acquires about a gradable social property is the ordering, because the ordering is what the training text contains. Manifesto percentages rarely appear in prose; relative claims about which party cares more about welfare are written down constantly. On that reading the model has no level to retrieve, and the upward shift we measure is what filling that gap looks like.

This complements \citet{lehmann2026shortcut}, who compare entity pairs on numerical attributes and find that the ordering implied by a model's own predicted values beats its direct comparison, meaning the values are held and the comparison underuses them. Running their comparison on our data produces no such gap, which is what the reading above predicts, because their magnitudes appear in text and can be routed around while ours never appear at all. It also sharpens what \citet{calderon2026recall} call the recall bottleneck, where facts are encoded but the key is lost and more inference-time effort recovers them. Here the ordering is reachable at more than twice baseline while the level stays below baseline however it is asked until it is supplied, which suggests the level is missing altogether and more effort would have nothing to recover.

Three results did not go our way, and each bounds the claim. Partial-credit scoring, introduced to be generous, made the comparison worse. Crediting adjacent buckets lifts placement to 58.8\%, but it lifts the constant-answer baseline to 80.7\%, so the models lose by a wider margin than under exact match. Our grounding condition is also weaker evidence than it first appears, for the reason given in the Limitations. Finally, one CMP category behaves opposite to its name and drew confident top-priority answers from every model, which is why we report it separately.

For evaluation the consequence is direct, and it may reach beyond politics. Benchmarks for gradable properties are overwhelmingly built from single-item questions, and if the asymmetry here generalises they measure the weaker capacity and report it as domain knowledge. Whether ``how severe'', ``how risky'' and ``how important'' questions share this shape is a hypothesis our two-country data cannot settle. The practical recommendation is to report both framings against their own baselines, since the two 58\% figures in Table~\ref{tab:dissociation} are indistinguishable until each is placed against what a constant answer would score.

The clearest next step is a finer scale. If placement fails because four buckets are too coarse, with the magnitude in fact learned, a seven- or eleven-point version of the same benchmark would show it, with the grounding condition supplying the ceiling to compare against. Beyond that, we would expect the dissociation wherever relative claims outnumber recorded quantities, so severity, risk and importance judgments are the natural tests of that hypothesis, and the agreement metric travels to them unchanged because it needs no answer key at all. Educational systems are a further setting. Multi-agent language-model systems already train teachers to manage hate incidents \citep{gajewska2025teachers} and tutor students through ethical debate \citep{maslowski2026debate}, and whether such tutors convey a position's level or only its ordering can be tested with the same two framings.

\section{Limitations}

The Manifesto Project's coding is itself imperfect: audits of coder agreement find real misclassification rates \citep{jankin2011coder}, and our gold labels inherit that uncertainty. The CMP measures salience rather than stance, so our claim concerns how much a party wrote about an issue, not what it said, and no result here is an assessment of any party. The self-consistency analysis rests on 80 comparable cases; the confidence intervals are reported and are wide, and the between-model comparison is underpowered.

The dissociation is measured on one property type in two countries, so the generalisation offered in the discussion is a hypothesis with a stated test, not a result. A finer ordinal scale might narrow the placement gap \citep{godfrey2025likert}, and we did not run the granularity sweep that would settle it. Our grounding result is weaker evidence against that objection than it may appear. It shows the four buckets can carry a correct answer, not that four buckets can faithfully express a graded belief the model holds ungrounded. What the objection does have to accommodate is that the ordering recovered from those same four-bucket answers is already as good as the directly-elicited one, so whatever resolution the coarse scale loses is not costing the ordinal signal. All prompting is in English, and cross-language discrepancies in political-information retrieval are documented \citep{kuai2025chatbot}. Our grounding hands the model the correct bucket outright instead of retrieving it through a realistic pipeline. One further model (Zhipu GLM-5) was queried but returned provider-side errors on every call and is excluded throughout.

\section{Conclusion}

Language models are increasingly asked where real actors stand, and such questions are gradable, carrying an order and a magnitude. Benchmarks for them are built almost entirely from single-item questions, so a model's score on one framing is read as its knowledge of the domain. Order and magnitude, however, need not be acquired together, because text about social entities carries far more comparison than measurement.

Using the Manifesto Project's coded data for parties in Japan and South Korea, we put seven production models to the same facts in two framings. Asked to order four parties they beat the best constant answer by a factor of two and a half; asked to place one party on the same scale they fell below it, under strict and partial-credit scoring alike. Supplying the coded values lifted placement to near ceiling on that same scale, consistent with a knowledge deficit rather than a comprehension one, and it significantly reduced the models' disagreement with their own answers on the lowest party, consistent with this instability following from the missing level. Along the way we found that how the verdict is read out of free prose moves the measured contradiction by twenty-eight points, which is a caution for reference-free metrics generally.

What a model knows about a gradable social property may therefore be the ordering rather than the level, and an evaluation that asks only for levels will understate the first and overstate its own reach. As these systems are increasingly consulted about what real parties and institutions stand for, which question is asked decides what the answer can mean.

\section*{Acknowledgments}
This work was partly supported by the VolkswagenStiftung under grant No.~98542 (project Deliberation Laboratory, DeLab).

\bibliography{references}

\appendix
\raggedbottom

\section{Related-Work Summary}
\label{app:related}

Table~\ref{tab:related} places this paper against the closest prior work by task, ground truth, and region.

\begin{table*}[h]
\centering
\small
\begin{tabular}{p{2.5cm}p{4.4cm}p{4.4cm}p{2.9cm}}
\toprule
\textbf{Work} & \textbf{Task} & \textbf{Ground truth} & \textbf{Source text given?} \\
\midrule
\citet{liusie2024comparative}, \citet{qin2023pairwise} & Judge quality of a supplied text, pairwise vs pointwise & Human preference & Yes \\
\citet{licht2025scalar} & Measure a scalar construct in a supplied text & Expert-coded scales & Yes \\
\citet{lehmann2026shortcut} & Compare two entities on a numerical attribute & Recorded attribute values & No (value is in text corpus) \\
\citet{calderon2026recall} & Recall an encoded fact & Web-grounded fact set & No \\
\citet{dileo2025mapping}, \citet{caliskan2025generative}, \citet{benoit2026salience} & Estimate party position or salience & CMP / expert surveys & Yes \\
This work & Order vs place a named party on a coded issue & CMP coded manifesto data & No \\
\bottomrule
\end{tabular}
\caption{The closest prior work by task, ground truth, and whether the text to be judged is supplied in the prompt. The last column is the axis this paper occupies alone: the model must answer from parametric knowledge, against an external key, in two framings.}
\label{tab:related}
\end{table*}

\section{Results by Issue Type}
\label{app:split}

\begin{table}[H]
\centering
\small
\setlength{\tabcolsep}{3pt}
\begin{tabular}{lccc}
\toprule
\textbf{Measure} & \textbf{All 9} & \textbf{Variant 6} & \textbf{Anchor 3} \\
\midrule
Order highest & 58.1 & 69.6 & 34.2 \\
\quad chance & 25.0 & 25.0 & 25.0 \\
Order lowest & 48.3 & 38.5 & 68.4 \\
\quad tie-aware chance & 35.3 & 25.0 & 56.6 \\
Place, exact & 22.4 & 21.9 & 23.6 \\
\quad best constant & 37.9 & 41.9 & 66.2 \\
Place, $\pm$1 & 58.8 & 63.2 & 49.7 \\
\quad best constant & 80.7 & 87.5 & 66.2 \\
\bottomrule
\end{tabular}
\caption{Headline measures split by issue type (\%, \texttt{per408} excluded). Ordering beats chance on the variant issues (highest $p\approx10^{-16}$, lowest $p=0.006$) but not on the anchor issues (highest $p=0.13$, lowest $p=0.094$), where all parties share a level; placement stays below its constant baseline in every subset.}
\label{tab:split}
\end{table}

\section{Multiple-Comparison Correction}
\label{app:holm}

\begin{table}[H]
\centering
\small
\setlength{\tabcolsep}{3pt}
\resizebox{\columnwidth}{!}{%
\begin{tabular}{lccc}
\toprule
\textbf{Test} & \textbf{$p$} & \textbf{Holm $\alpha$} & \textbf{Holds} \\
\midrule
Calibration across models, $\chi^2$ & $<10^{-24}$ & 0.0056 & yes \\
Order highest, binomial & $3.0\times10^{-14}$ & 0.0063 & yes \\
Read-out SC highest, Fisher & 0.0001 & 0.0071 & yes \\
Grounding SC lowest, Fisher & 0.0006 & 0.0083 & yes \\
Order lowest, binomial (tie-aware) & 0.0028 & 0.0100 & yes \\
Grounding SC highest, Fisher & 0.039 & 0.0125 & no \\
Read-out SC lowest, Fisher & 0.063 & 0.0167 & no \\
Between-model SC, $\chi^2$ & 0.231 & 0.025 & no \\
Implied vs direct ordering, Fisher & 0.54 & 0.05 & no \\
\bottomrule
\end{tabular}}
\caption{Holm correction over the nine tests reported in Section~\ref{sec:results}. The last row is a null result by design: the implied and direct orderings are expected to agree.}
\label{tab:holm}
\end{table}

\section{Evaluation Procedure}
\label{app:algo}

Algorithm~\ref{alg:eval} states the full pipeline. Steps 1--7 produce the placement and ordering accuracies of Section~\ref{sec:results} and their constant-answer baselines; steps 8--10 compute the key-free self-consistency measure (Eq.~\ref{eq:sc}); step 11 is the grounding intervention, which reruns the pipeline with the coded facts prepended.

\begin{algorithm}[H]
\caption{Ordering vs Placement Evaluation}
\label{alg:eval}
\begin{algorithmic}[1]
\Require parties $P$, issues $J$, models $M$, CMP percentages $c_{p,j}$
\Ensure placement accuracy, ordering accuracy, their baselines, $\mathrm{SC}^{\mathrm{hi}}, \mathrm{SC}^{\mathrm{lo}}$
\For{$m \in M,\ p \in P,\ j \in J$}
 \State query absolute prompt; parse tagged bucket $\hat{b}^{\mathrm{abs}}_{m,p,j}$
\EndFor
\For{$m \in M$, country, $j \in J$}
 \State query comparative prompt; parse tagged $r^{\mathrm{hi}}_{m,j}, r^{\mathrm{lo}}_{m,j}$
\EndFor
\For{each absolute item}
 \State score vs $b(c_{p,j})$: correct / partial / rival / none / no\_knowledge
\EndFor
\State baselines $\gets$ best constant answer under each scoring rule
\For{$(m,j) \in \mathcal{C}$ (comparable cases)}
 \State $s_{m,j} \gets \arg\max_p \hat{b}^{\mathrm{abs}}_{m,p,j}$; agreement $\gets \mathbb{1}[s_{m,j} \in \tau(r^{\mathrm{hi}}_{m,j})]$
\EndFor
\State $\mathrm{SC}^{\mathrm{hi}} \gets$ mean agreement (Eq.~\ref{eq:sc}); $\mathrm{SC}^{\mathrm{lo}}$ analogous
\State grounding: repeat with coded facts prepended; compare via Fisher's exact
\State \Return accuracies, baselines, $\mathrm{SC}^{\mathrm{hi}}, \mathrm{SC}^{\mathrm{lo}}$, grounded $\mathrm{SC}$
\end{algorithmic}
\end{algorithm}

\end{document}